\documentclass{article}

\usepackage{arxiv}

\usepackage[utf8]{inputenc} 
\usepackage[T1]{fontenc}    
\usepackage{hyperref}       
\usepackage{url}            
\usepackage{booktabs}       
\usepackage{amsfonts}       
\usepackage{nicefrac}       
\usepackage{microtype}      
\usepackage{lipsum}
\usepackage{graphicx}
\graphicspath{ {./images/} }
\usepackage{amsmath}
\usepackage{url}
\usepackage{threeparttable}

\title{Improving Landslide Detection on SAR Data through Deep Learning}

\author{
 Lorenzo Nava \\
  Department of Earth Sciences\\
  University of Florence\\
  Florence 50121, Italy \\
  \texttt{lorenzo.nava@stud.unifi.it} \\
   \And
 Oriol Monserrat \\
  Department of Remote Sensing\\
  Centre Tecnològic Telecomunicacions Catalunya\\
  Castelldefels 08860, Spain \\
  \texttt{omonserrat@cttc.cat} \\
  \And
 Filippo Catani \\
  Department of Geosciences\\
  University of Padua\\
  Padua 35131, Italy \\
  \texttt{filippo.catani@unipd.it} \\
}

\begin{document}
\maketitle
\begin{abstract}
In this letter, we use deep-learning convolution neural networks (CNNs) to assess the landslide mapping and classification performances on optical images (from Sentinel-2) and SAR images (from Sentinel-1). The training and test zones used to independently evaluate the performance of the CNNs on different datasets are located in the eastern Iburi subprefecture in Hokkaido, where, at 03.08 local time (JST) on September 6, 2018, an Mw 6.6 earthquake triggered about 8000 coseismic landslides. We analyzed the conditions before and after the earthquake exploiting multi-polarization SAR as well as optical data by means of a CNN implemented in TensorFlow that points out the locations where the Landslide class is predicted as more likely. As expected, the CNN run on optical images proved itself excellent for the landslide detection task, achieving an overall accuracy of 99.20\% while CNNs based on the combination of ground range detected (GRD) SAR data reached overall accuracies beyond 94\%. Our findings show that the integrated use of SAR data may also allow for rapid mapping even during storms and under dense cloud cover and seems to provide comparable accuracy to classical optical change detection in landslide recognition and mapping.
\end{abstract}


\section{Introduction}

In the world, several natural phenomena, like earthquakes and intense rainfalls, sometimes combined with windstorms, can trigger multiple landslide events, that can occur in groups of hundreds or even thousands in a region [\ref{1}-\ref{5}]. Those events can cause noticeable damages to natural and human infrastructures and can cause physical damages and heavy economic and social impacts [\ref{6}]. Therefore, there is a growing need to intervene quickly and efficiently in the affected areas. Numerous techniques have been elaborate to produce susceptibility maps [\ref{7}], [\ref{8}], and mitigation strategies [\ref{9}], [\ref{10}]. Undoubtedly, in the case of multiple occurrences of landslides over a large area, the most common mapping method relies on remote sensing data because of its potential to quickly mapping geological features of whole regions, without physical contact with the areas investigated. A vast amount of research exists that has been performed to this end with knowledge-based methods, multiple regression, analytic hierarchy process [\ref{11}], and machine learning (ML) techniques [\ref{12}-\ref{15}]. However, as time is a key factor when mapping natural disaster effects [\ref{16}], usage of optical data alone can have serious limitations in presence of cloud cover. A possible solution, offered by the usage of SAR satellites, has been exploited only scantly, so far, because landslide mapping algorithms have been mainly developed to run on optical data together with ancillary geological and topographic stationary information and because SAR data require specific analysis methods and classification algorithms, not yet fully developed for mass movements on natural slopes. Therefore, a research gap is still present in the lack of reliable image analysis methods to extract landslide mapping information from SAR images. Furthermore, it is not available, in the bibliography, an automatic method that employs the combination of CNNs and SAR data to extract spatial landslide information, probably due to the characteristics of the SAR data, which provides different information compared to the mainly exploited optical data, making essential the development of innovative techniques. Recently, deep learning (DL) approaches and, mainly, convolutional neural networks (CNNs) have been used in various remote sensing tasks on VHR imagery, such as classification, segmentation [\ref{17}], and object detection [\ref{18}]. Nevertheless, few are the studies that use CNNs for landslide detection. Ding et al. [\ref{19}] carried out an automatic recognition of landslide at pixel scale based on CNNs on GF-1 images with four spectral bands (blue, green, red, and near-infrared) and a spatial resolution of 8 m, achieving a detection rate of 72.5\%, a false positives rate of 10.2\%, and overall accuracy of 67\%. Ghorbanzadeh et al. [\ref{20}] evaluated various ML algorithms on VHR optical data from the Rapid Wye satellite and topographic factors achieving 78.26\% mIOU for a small window-size CNN. Another interesting study was carried out by Catani [\ref{21}], in which the author evaluates different state-of-art CNNs on non-nadiral and crowdsourced optical images of landslides to classify them, achieving overall accuracies between 87\% and 90\%. Indeed, optical images are great tools, but they present limitations due to the cloud cover that, in regions as the Congo River basin, Equatorial South America, and Southeast Asia, can display annual cloud frequencies higher than 80\% [\ref{22}]. In some cases, the first cloudless image available in the service after multiple landslide events occurred about a month later, as happened in Chile in December 2017, in Nepal in 2015 [\ref{23}], and in Ecuador in 2016. Despite numerous methods and techniques have been applied on SAR data to detect landslides, as can be seen in the review of Mondini et al. [\ref{24}], our study is the first that provides, to this end, a DL-based method employing the combination of CNNs and SAR data. Furthermore, the accuracy achieved is comparable to optical change detection. 

In the remainder of this letter,  we show and discuss a  deep-learning-based method to detect landslides also in case of illumination or atmospheric conditions not favorable for landform mapping. We used convolutional neural networks  (CNNs)  on various combinations of Sentinel-1 data and topographic factors. We apply CNN methods to landslide detection based on ground range detected (GRD) SAR imagery from the Sentinel-1 satellite, which, in a study carried out by Mondini et al. [20], showed unambiguous changes of amplitude caused by landslides in about eighty-four percent of the cases. Lastly, we compare the results obtained on eight different SAR-based datasets with an RGB dataset, and we provide the mapping of the study area by using the best models.

\section{Study Area and Materials}
\paragraph{Study Area}
Our case study area lies in the Iburi sub-prefecture of Hokkaido, Japan. It comprehends the Atsuma, Mukawa, and Abira towns, which present a specific population of less than 10,000 and a low population density of 17 inhabitants/km². The morphology of the area is composed mainly of hills. The maximum height is fewer than 800 m while the average elevation is 160 m. The basement complex of the region is mainly composed of sedimentary rocks of the Neogene tertiary system: layers of sandstone and mudstone, sandstone, conglomerate, and diatomaceous siltstone [\ref{26}], [\ref{27}]. In the study area, at 03.08 local time (JST) on September 6, 2018, struck an Mw 6.56 earthquake (HEIE). It was activated by a rupture of a low-activity blind fault with the epicenter located at 42.690 N 142.007 E [\ref{28}]. The event triggered about 7837 coseismic landslides most of which occurred in locations where the elevation is less than 300 m [\ref{4}]. The majority of the landslides slid down over the air-fall pumice and ash layers are shallow and are classified as planar type and spoon type [\ref{30}]. 

\paragraph{Materials}
The Sentinel-1 mission encompasses two polar-orbiting satellites that perform C-band synthetic aperture radar imaging [\ref{31}] while the Copernicus Sentinel-2 mission comprises two polar-orbiting satellites that sample 13 spectral bands: four bands at 10 m, six bands at 20 m, and three bands at 60 m spatial resolution [\ref{32}]. We downloaded Sentinel-1 and Sentinel-2 images from the Copernicus Open Access Hub [\ref{33}]. The first images (5 x 20 m) were acquired in Level-1 Ground Range Detected (GRD) mode, with VV and VH polarization, and Interferometric Wide (IW) acquisition mode. GRD products are focused SAR data that has been detected, multi-looked, and projected to ground range using an Earth ellipsoid model [\ref{34}]. Images were acquired from three different days: a) 01 September 2018, b) 13 September 2018, c) 25 September 2018. Lastly, a Sentinel-2 VHR RGB image (10 x 10 m) was obtained from the first cloud-free day after the multiple landslide event (20 October 2018). 

\begin{table}[!th]

\renewcommand{\arraystretch}{1.3}
\setlength{\tabcolsep}{40pt}
\label{tab:S1-2data}
\centering
\begin{threeparttable} \caption{Dates of Acquisition of Sentinel-2 and Sentinel-1
for the Study Area in Hokkaido in 2018.}
\begin{tabular}{ccc}
\hline
\textbf{Region and Year}      & \textbf{Sentinel-2} & \textbf{Sentinel-1} \\ \hline
                     &            & 01-09      \\
Iburi, Hokkaido 2018 & 20-10      & 13-09      \\
                     &            & 25-09      \\ \hline
\end{tabular}
\end{threeparttable}
\end{table}

Surface topography is one of the most influential elements regarding landslides in hilly and mountainous areas and, above all, the slope angle is considered an essential component of slope stability analysis [\ref{35}]. Therefore, in this study, we used a 30 m resolution digital elevation model (DEM) acquired from USGS Earth Explorer [\ref{36}] and slope angle to evaluate their impact on the SAR data. As quoted, we designed an RGB dataset and eight SAR-based datasets, each composed of different combinations of the data in Table \ref{tab:S1-2data}, as described in Table \ref{tab:Comp-images}.

\begin{table}[!th]
\renewcommand{\arraystretch}{1.3}
\setlength{\tabcolsep}{20pt}
\label{tab:Comp-images}
\centering
\begin{threeparttable}\caption{Composite images created with corresponding dataset name.}
\begin{tabular}{cccc}
\hline
\textbf{Dataset Name}      & \textbf{Band1} & \textbf{Band2} & \textbf{Band3}\\ \hline
RGB                  &     Red       & Green     &  Blue\\
SSD                  & VV after-event      & VH after-event     & DEM\\
SSS                  & VV after-event      & VH after-event     & Slope\\
BAD                  & VV before-event      & VV after-event     & DEM\\
BAS                  & VV before-event      & VV after-event     & Slope\\
HHH                  & VH before-event      & VH after-event     & VH after-event\\
BAA                   & VV before-event      & VV after-event     & VV after-event\\
BAC                  & VV before-event      & VV after-event     & \begin{tabular}[c]{@{}c@{}}VV after-event -\\  VV before-event\end{tabular} \\
BAH                   & VV before-event      & VV after-event     & VH after-event\\ \hline
\end{tabular}
\end{threeparttable}
\end{table}

\section{Methodology}

The radar information from Sentinel-1 is exploited in combination with topographic factors, to evaluate the performance of a CNN to detect landslides. 
The workflow of this study is as follow:
\begin{itemize}
    \item Designing nine different datasets, one composed of RGB images, four composed by SAR data and a topographic factor as third band, of which two with slope and two with DEM, and four constituted only by SAR data.
    \item Training the CNN on each training dataset and validating the performance on the corresponding test dataset sampled in the study area.
    \item Visualizing the predictions of the best models on the entire study area.
\end{itemize}
In the following sections, we give a description and the results of this workflow. Additional information and considerations are provided in the conclusion section.

\subsection{Dataset Pool}

To create the datasets, each of the composite images in Table \ref{tab:Comp-images} is sampled in the GIS environment using the '\textit{Export training data for deep learning}' tool, that, after designing numerous polygons on known landslides, samples various clips per polygon with a predefined shape, stride, and resolution. The polygons position is chosen concerning the high variability of both classes, being careful to include various landslide shapes, orientations, and dimensions. Besides, we include different land covers, such as urban, wooded, and agricultural areas in the \textit{Non-landslide} class. In all the datasets, the spatial resolution of the images is 25 x 25 pixels, which is equivalent to a spatial extension of 250 x 250 m for optical RGB images and 125 x 420 m for SAR products. We select this image size as the best size based on a cross-validation for our study area.

\subsection{Supervised Classification}

\begin{figure}[!th]
\centering
\includegraphics[scale=0.41]{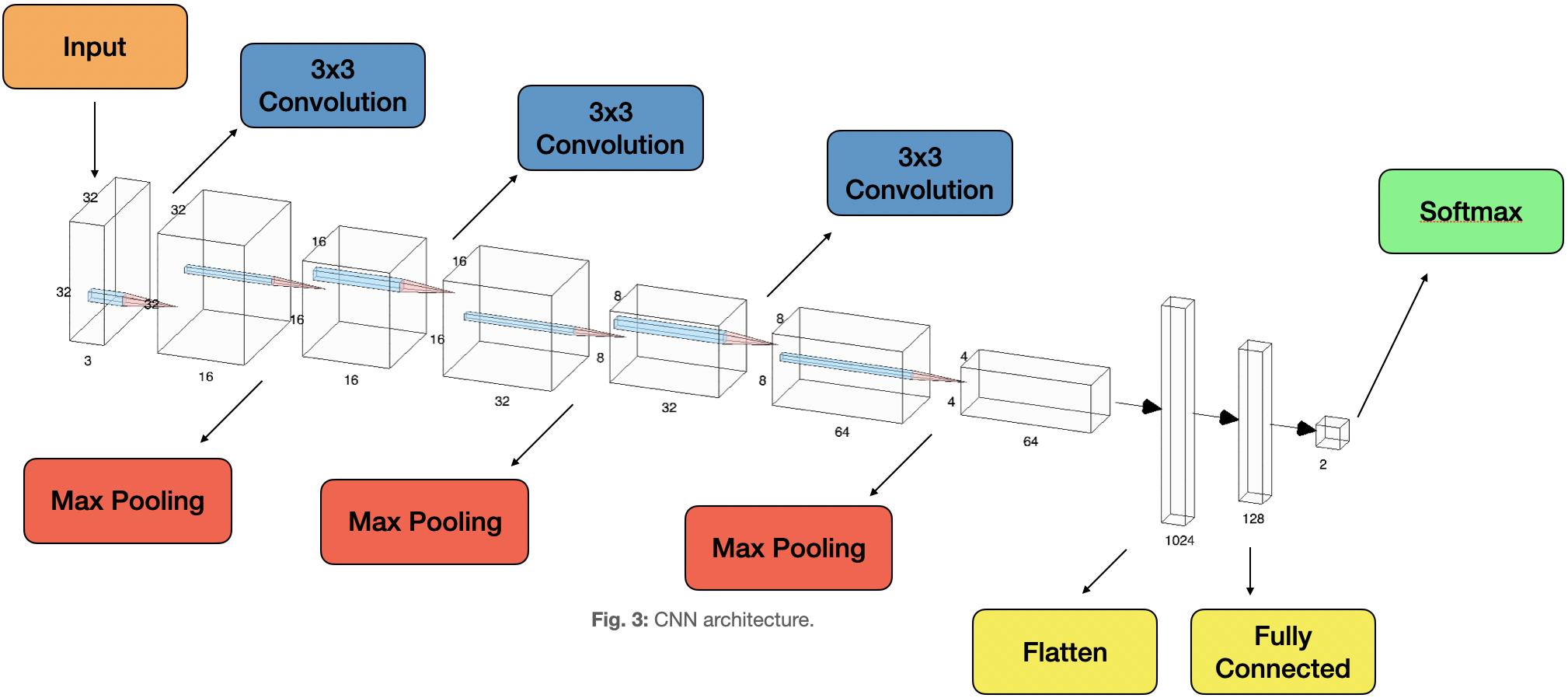}
\caption{CNN architecture.}
\label{fig:mycnn}
\end{figure}

Recently, deep-learning methods and, above all, CNNs have accomplished reliable results in various tasks in computer vision, proving to be the state-of-art of this field [\ref{37}], [\ref{38}]. A CNN can autonomously learn hierarchical feature representations of an image, enabling it to perform classification tasks directly from images, e.g., recognizing a specific image without using human-designed features [\ref{39}]. The peculiarity of a CNN is its architecture, which is composed of tens or hundreds of so-called hidden layers that exercise convolutional and pooling operations. During the training process, an input image flows through the network that scans it with a set of trainable kernels, resulting in a group of feature maps (forward phase), then gradients are back-propagated, and parameters are updated (backward phase). The output of each convolutional layer is filtered by an activation function (e.g., sigmoid, ReLU, Softmax, hyperbolic tangent) that performs non-linear transformations [\ref{40}]. The pooling layer, usually placed after the convolutional layer, aims to simplify the output by performing non-linear down-sampling to reduce the parameters. 

In this letter, we modify a network provided by the TensorFlow Core \textit{Image Classification} library [\ref{41}]. The model consists of three convolution blocks of 3x3 filters with a max pooling layer in each of them. Dropout and connected layer are activated by a ReLU activation function: \begin{equation}R(z)=max(0, z)\end{equation}
The last fully connected layer has 2 units, as the number of classes we defined, and it is activated by a Softmax activation function:
\begin{equation} \sigma(x)_j=\dfrac{e^{z_j}}{\begin{matrix} \sum_{k=1}^K e^{z_k} \end{matrix}} \textrm{ for }j=1,\dots, K \textrm{ and } z \in R\end{equation} 
As loss function optimization, we used a stochastic gradient descent method based on an adaptive estimation of first-order and second-order moments (Adam), a method for problems with very noisy and/or sparse gradients [\ref{42}]. Sparse Categorical Crossentropy (SCC) function is used as a loss function, which is an integer-based version of the Categorical Crossentropy function:
\begin{equation}Loss=- \begin{matrix} \sum_{i=1}^M y_i\cdot\log \hat{y}_i\end{matrix}\end{equation} where \begin{math} \hat{y}_i\end{math} is the \textit{i}-th scalar value in the model output, \begin{math} y_i\end{math} is the corresponding target value, and M, the output size is the number of scalar values in the model output. The SCC function computes the cross-entropy loss between the labels and predictions providing labels as integers [\ref{43}]. The CNN is implemented using the Google’s library TensorFlow [\ref{44}]. Furthermore, many augmentation techniques are applied on the dataset, such as vertical and horizontal \textit{Random Flip}, \textit{Random Rotation}, \textit{Random Zoom}, and \textit{Random Translation}. Those are chosen considering that flipping, rotating, zooming, and translating a satellite landslide image results in new images of landslides with a possible real dimension and orientation. The trained model is used to predict classes on a labeled test dataset composed of images with the same number and type of bands of the training dataset. The output contains two classes, and accuracy (4), precision (5), and recall (6) are calculated using these values according to the following formulas, where TP, TN, FP, and FN are true and false positives and negatives, respectively.

\begin{equation}
\textrm{Accuracy = } \dfrac{TP + TN}{TP + FP + TN + FN}
\end{equation}

\begin{equation}
\textrm{Precision = } \dfrac{TP}{TP + FP}
\end{equation}

\begin{equation}
\textrm{Recall = } \dfrac{TP}{TP + FN}
\end{equation}

\subsection{Object Detection}
\label{subsec:Obj}
The study area is analyzed through the sliding window [\ref{45}] method to detect landslides. The maps created are shown with superimposed the shapefile of the pre-mapped landslides (yellow) and a red point in correspondence to the center of the patches predicted as \textit{Landslide}. The step of the window is 2 pixels. Patches are extracted with 25 x 25 pixels resolution and then resampled to match the input size of the network (\textit{n} x 32 x 32 x 3).

All experiments were executed on a mac-OS operating system computer with a 2.2 GHz Intel Core i7 with 6 cores, a 256 GB SSD for quick access to applications and datasets, and RAM 16 Gb.

\section{Results}
\subsection{Landslide Classification}

\begin{table}[!th]
\renewcommand{\arraystretch}{1.3}
\setlength{\tabcolsep}{25pt}
\label{tab:results}
\centering
\begin{threeparttable}
\caption{Accuracy, precision and recall of the CNN tested on eight different test datasets explained in Table \ref{tab:Comp-images}. The best results achieved with SAR data are pointed out.}
\begin{tabular}{cccc}
\hline
\textbf{Dataset Name} & \textbf{Accuracy(\%)} & \textbf{Precision(\%)} & \textbf{Recall(\%)} \\ \hline
RGB                   & 99.20                 & 99.60                  & 98.81               \\ \cline{2-4} 
SSD                   & 86.63                 & 92.24                  & 80.16               \\
SSS                   & 80.84                 & 82.77                  & 78.17               \\
BAD                   & 89.96                 & 95.83                  & 83.47               \\
BAS                   & 88.55                 & 96.58                  & 79.84               \\
HHH                   & 81.87                 & 93.22                  & 68.75               \\
BAA                   & \textbf{94.17}                 & \textbf{97.32}                  & 90.83               \\
BAC                   & 90.83                 & 93.36                  & 87.92               \\
BAH                   & 93.33                 & 95.22                  & \textbf{91.25} \\ \hline
\end{tabular}
\end{threeparttable}
\end{table}

We trained each model for the optimal number of epochs and using the best learning rate that fitted each specific problem. Table \ref{tab:results} shows the results for different datasets. As expected, the model trained and tested on RGB optical images performs well, achieving the best results. Focusing on SAR datasets, the BAH and BAA achieved the highest overall accuracies, with 93.33\% and 94.17\%, respectively. Besides, the BAH, achieved also the highest recall, with a value of 91.25\%. The two datasets enclosing the slope as a third band, SSS and BAS, generally achieved the lowest accuracies, with 80.84\% and 88.55\%, respectively. Lastly, considering the HHH and the BAA, which are composed of the same combination of bands but different polarization, results show that the VV polarization is more discriminating to detect landslides than the VH.

\begin{figure}[!t]
\centering
\includegraphics[scale=0.60]{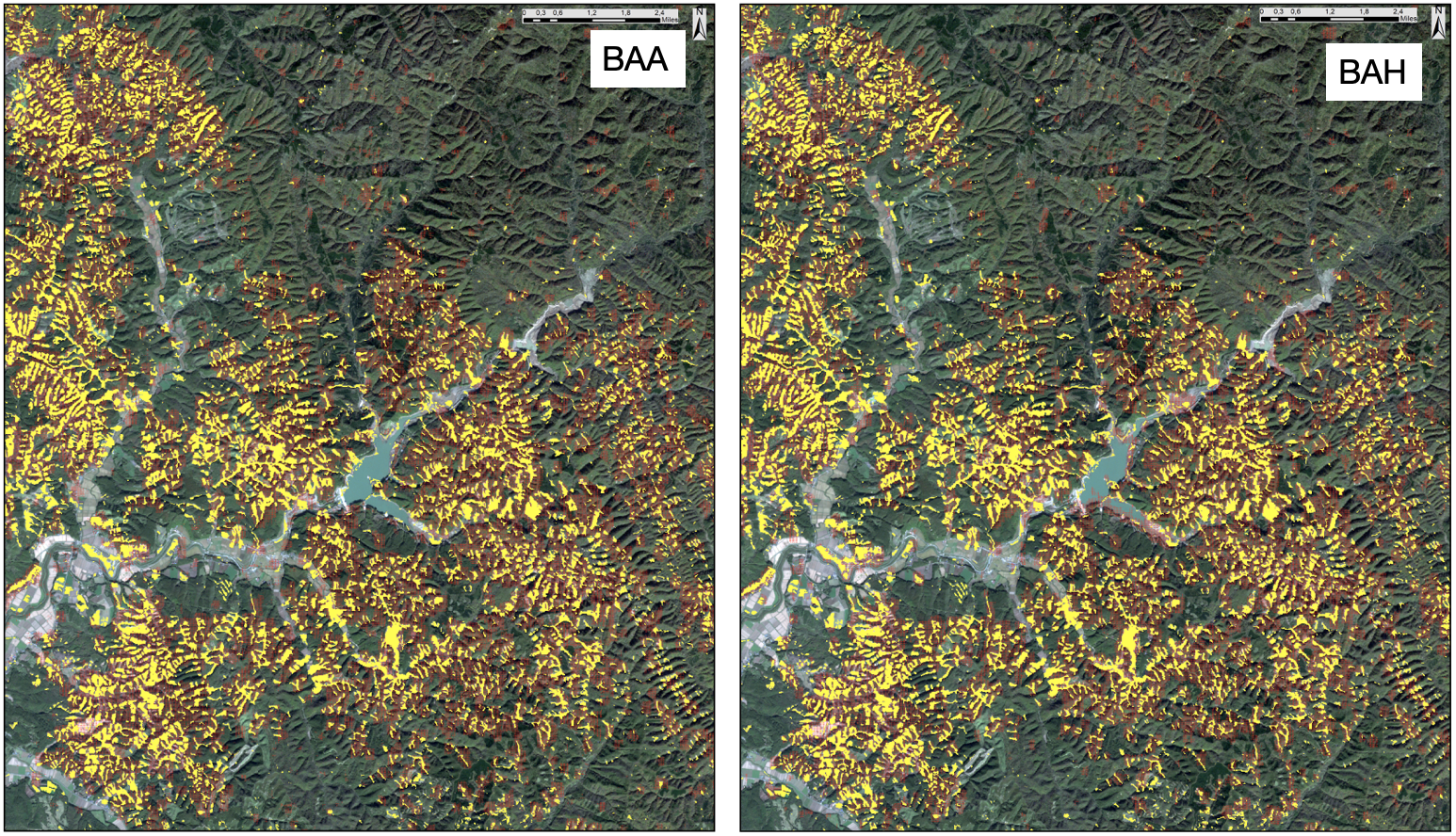}
\caption{Predictions of the entire study area by the models trained on the BAA and BAH datasets. (BAA: Accuracy = 94.17\%; Precision = 97.32\%; Recall = 90.83\%. BAH: Accuracy = 93.33\%; Precision = 95.22\%; Recall = 91.25\%.)}
\label{fig:BAHmap}
\end{figure}

To accomplish a visual evaluation of the predictions of the best models trained in this study, we analyzed the entire study area through the process explained in Section \ref{subsec:Obj}. Figure \ref{fig:BAHmap} shows the resulting mapping obtained with the models trained on the BAA and BAH datasets.

\section{Conclusion}
In this letter, we analyzed and discussed various combinations of SAR imagery and topographic factors and we proposed a DL-based method for landslide detection to locate landslides in a short time, also during the night and in presence of cloud cover. The best overall accuracy was accomplished by the BAA and BAH datasets, composed just by three SAR bands. The accuracy of the latter was 94.17\% and 93.33\%, respectively, 5.03\% and 5.87\% less than with the RGB dataset, but with the advantage of being applicable also during storms or night. Therefore, it is possible to make a landslide detection without VHR optical images with similar accuracy values, using the combination of SAR data of the BAA or the BAH datasets. Moreover, VV and VH amplitude imagery from Sentinel-1 is an open-source product available in almost all parts of the globe. The results achieved so far in this letter are promising and suggest that SAR Sentinel-1 images and deep learning models are a trustworthy combination for locating rapid landslides. SAR images permit us to obtain information regarding landslides also during a rain event or during the night when optical imagery, instead, is unusable or not available because of the presence of cloud cover. Therefore, the method could involve also various benefits in terms of emergency management civil protection operations, such as significantly decreasing the time of the emergency response in various emergency scenarios by increasing the quality of hazard mapping and risk assessments.

\newpage

\bibliographystyle{unsrt}  


\end{document}